\documentclass[conference]{IEEEtran}
\IEEEoverridecommandlockouts
\usepackage{tabularx}
\usepackage{cite}
\usepackage[most]{tcolorbox}
\usepackage{amsmath,amssymb,amsfonts}
\usepackage{algorithmic}
\usepackage{graphicx}
\usepackage{textcomp}
\usepackage{xcolor}
\usepackage{caption}
\usepackage{booktabs} 
\usepackage{comment}
\usepackage[shortcuts,acronym]{glossaries}
\usepackage[hidelinks]{hyperref}
\usepackage{enumitem}
\usepackage{float}
\usepackage{stfloats}

\def\BibTeX{{\rm B\kern-.05em{\sc i\kern-.025em b}\kern-.08em
    T\kern-.1667em\lower.7ex\hbox{E}\kern-.125emX}}

\begin{document}

\title{Empathy by Design: Aligning Large Language Models for Healthcare Dialogue}

\author{
\IEEEauthorblockN{
Emre Umucu\IEEEauthorrefmark{1}\textsuperscript{*},
Guillermina Solis\IEEEauthorrefmark{3},
Leon Garza\IEEEauthorrefmark{2},
Emilia Rivas\IEEEauthorrefmark{2},
Beatrice Lee\IEEEauthorrefmark{4},
Anantaa Kotal\IEEEauthorrefmark{2},
Aritran Piplai\IEEEauthorrefmark{2}
}

\IEEEauthorblockA{
\IEEEauthorrefmark{1}Department of Public Health Sciences, The University of Texas at El Paso, USA\\
Email: eumucu@utep.edu
}

\IEEEauthorblockA{
\IEEEauthorrefmark{3}Department of Nursing, The University of Texas at El Paso, USA\\
Email: gsolis2@utep.edu
}

\IEEEauthorblockA{
\IEEEauthorrefmark{2}Department of Computer Science, The University of Texas at El Paso, USA\\
Emails: \{lgarza3, erivas6\}@miners.utep.edu, \{akotal, apiplai\}@utep.edu
}

\IEEEauthorblockA{
\IEEEauthorrefmark{4}Department of Rehabilitation Sciences, The University of Texas at El Paso, USA\\
Email: ylee6@utep.edu
}

\thanks{\textsuperscript{*} Corresponding authors}
}


\newacronym{llm}{LLM}{Large Language Model}
\newacronym{llms}{LLMs}{Large Language Models}
\newacronym{dpo}{DPO}{Direct Preference Optimization}

\maketitle

\begin{abstract}
General-purpose large language models (LLMs) have demonstrated remarkable generative and reasoning capabilities but remain limited in healthcare and caregiving applications due to two key deficiencies: factual unreliability and a lack of empathetic communication. These shortcomings pose significant risks in sensitive contexts where users, particularly non-professionals and caregivers, seek medically relevant guidance or emotional reassurance. To address these challenges, we introduce a Direct Preference Optimization (DPO)-based alignment framework designed to improve factual correctness, semantic coherence, and human-centric qualities such as empathy, politeness, and simplicity in caregiver–patient dialogues. Our approach fine-tunes domain-adapted ~\ac{llms} using pairwise preference data, where preferred responses reflect supportive and accessible communication styles while rejected ones represent prescriptive or overly technical tones. This direct optimization method aligns model outputs with human preferences more efficiently than traditional reinforcement-learning-based alignment. Empirical evaluations across multiple open and proprietary LLMs show that our DPO-tuned models achieve higher semantic alignment, improved factual accuracy, and stronger human-centric evaluation scores compared to baseline and commercial alternatives such as Google’s medical dialogue systems. These improvements demonstrate that preference-based alignment offers a scalable and transparent pathway toward developing trustworthy, empathetic, and clinically informed AI assistants for caregiver and healthcare communication.  Our open-source code is accessible at: https://github.com/LeonG19/Empathy-by-Design

\end{abstract}

\begin{IEEEkeywords}
healthcare communication, large language models, empathy in AI, caregiver support, human-centered artificial intelligence
\end{IEEEkeywords}


\section{Introduction}
\label{sec:intro}

Caring for individuals with chronic or neuro-degenerative conditions such as Alzheimer’s disease and dementia requires not only clinical coordination but also constant emotional resilience. Family caregivers and care partners often become the primary interpreters of medical information, navigating complex treatment decisions, behavioral changes, and communication challenges on a daily basis. While they cultivate deep experiential knowledge of the individuals in their care, they face uncertainty when medical guidance is inaccessible or overly technical. In such moments, caregivers increasingly turn to online resources and, notably, to ~\ac{llms} for immediate, conversational support. \ac{llms} have rapidly become integrated into everyday life. They can explain complex ideas in plain language, adjust to a user’s tone, and offer a sense of understanding that static websites cannot. For caregivers seeking clear, kind, and quick answers, these systems can feel like an always-available companion in moments of doubt or stress.

However, the growing reliance on generative models raises important concerns and potential risks\cite{yankouskaya2025can}; particularly in healthcare. Artificial Intelligence (AI) is increasingly embedded in clinical practice, as seen in the number of FDA-approved medical AI devices rising from just 2 per year in 2016, to 69 in 2022 \cite{pal2025generative, fda2025ai_enabled_medical_devices}. These tools undergo rigorous validation procedures to ensure accuracy, are deployed by healthcare professionals, and are continually monitored and reported back to the FDA. In professional settings, such risks are carefully managed. Outside clinical settings, however, a growing number of individuals, such as caregivers, patients, and family members, are turning to online resources powered by \ac{llms} to simplify complex terms, explore treatment options, and, in some cases, even attempt self-diagnosis \cite{deBronkart2024three_years}. The challenge arises when these unregulated models are used for medical guidance: without clinical oversight, they can produce inaccurate, incomplete, or misleading information that influences real-world health decisions \cite{grasso2025patients}. Thus the question remains, what happens when non-professionals, such as caregivers, family members, or patients with chronic conditions like dementia or Alzheimer’s, begin relying on tools that have not been FDA-approved for their healthcare questions? 

\ac{llms} are trained on vast amounts of data, much of which has not been clinically validated \cite{chouffani2024ai_validation}. While this may seem minor, the risk is significant: an \ac{llm} could generate inaccurate, incomplete, or confusing responses, which in turn may lead to misdiagnosis, oversimplification of complex issues, or the spread of misinformation \cite{pal2025generative}. In a study conducted by Nielsen et al. \cite{nielsen2023validity}, a group of doctors evaluated GPT-4's responses to medical questions using the Likert scale, assessing the accuracy, relevance, and depth. They concluded that GPT-4's answers lacked depth and were concerned that because it was trained on publicly available text, its answers would potentially disseminate biases. In another study by Zada et al. \cite{zada2025medical}, \ac{llms} were found to be correct in only 31\% of scenarios when asked open-ended questions, mimicking real-world self-diagnosis use cases. The authors concluded that there is a critical need for a comprehensive self-diagnosis dataset to improve the performance of current \ac{llms}, with the goal of enhancing their reliability and potentially enabling their integration into future healthcare systems. 

\begin{figure*}[t]
    \centering
    \includegraphics[width=.95\textwidth, height=0.70\textheight, keepaspectratio]{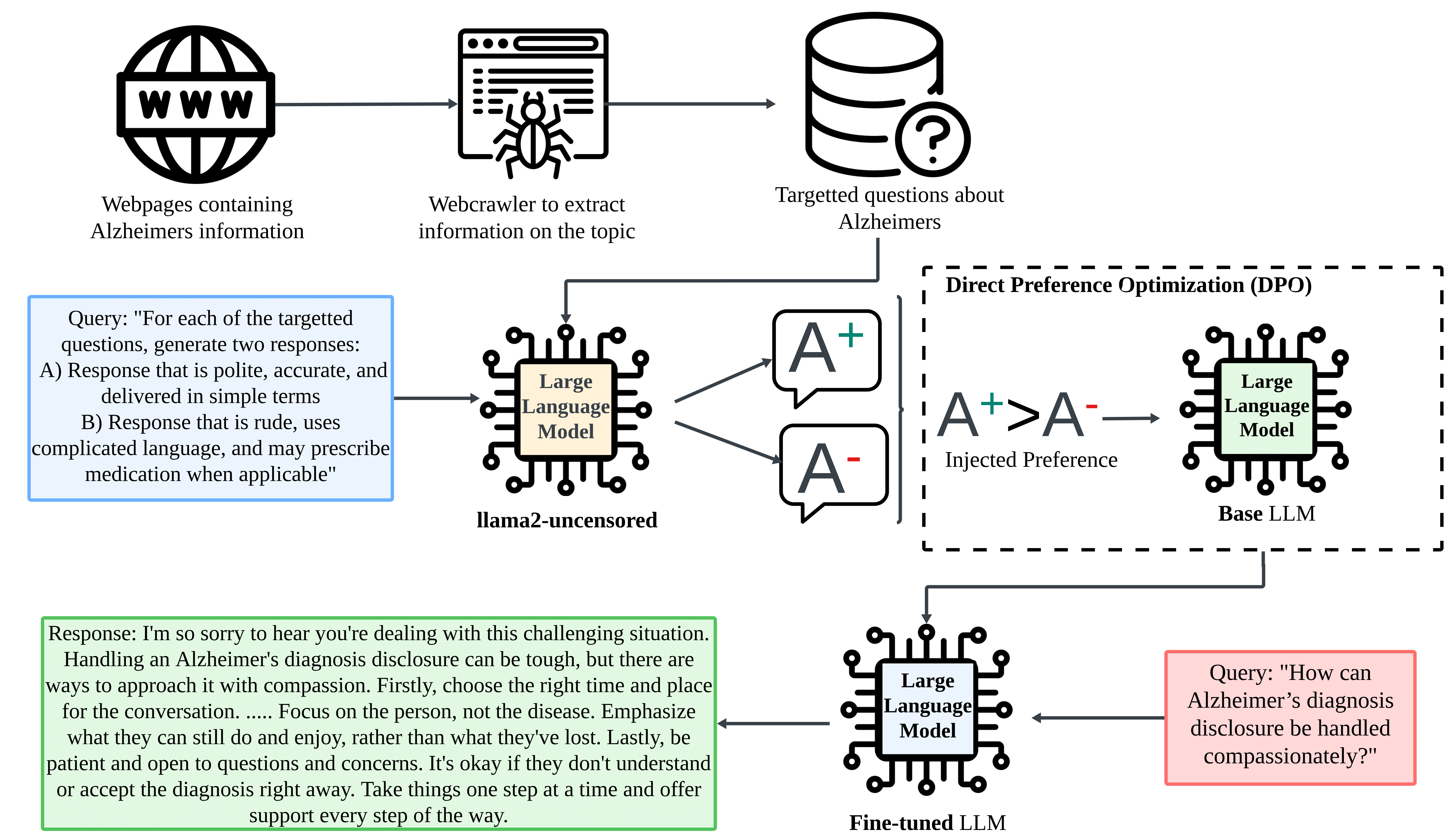}
    \caption{Overview of the proposed DPO fine-tuning workflow for healthcare language models. Web-scraped Alzheimer’s resources are transformed into paired question–answer examples, where preferred (A$^{+}$) responses emphasize empathy and factual accuracy, and rejected (A$^{-}$) responses contain undesirable traits. Direct Preference Optimization (DPO) aligns the base model toward the preferred behavior, producing a fine-tuned LLM capable of compassionate and reliable caregiving dialogue.}

    \label{fig:framework_architecture}
\end{figure*}

Beyond factual accuracy, empathy plays a vital role in effective healthcare communication. Research consistently shows that empathetic communication improves patient trust, adherence, and emotional outcomes, particularly in long-term or memory-related conditions such as dementia \cite{deist2015resilience}. Caregivers often seek not only correct information but also reassurance and emotional validation, elements that strongly influence how they interpret and act on medical guidance \cite{palacio2020resilience}. When language models respond in a cold or overly clinical tone, users may perceive them as dismissive or judgmental, which can increase anxiety and reduce engagement \cite{srinivasan2022role, chi2023investigating}. Conversely, responses that acknowledge emotional context and convey understanding foster a sense of support and reliability, even when the advice remains non-prescriptive \cite{sorin2024large}. Therefore, an effective healthcare-oriented \ac{llm} must balance factual precision with empathetic communication, ensuring that its guidance is both trustworthy and emotionally attuned to the needs of caregivers and patients.

Taken together, these concerns point to a broader challenge: how can large language models communicate health-related information in ways that are both factually reliable and emotionally supportive? Our goal is to design a model that delivers clear, unbiased, and easy-to-understand responses in a supportive and friendly tone, supplying patients and caregivers access to trustworthy guidance without over-complication or negativity, while avoiding prescriptive medical advice.

This paper explores whether preference-based fine-tuning can bridge this gap. Specifically, we investigate whether aligning models using \ac{dpo} can enhance both the factual reliability and empathetic tone of responses in healthcare-oriented dialogue. By focusing on how models express information, not just what they say, we aim to build systems that deliver accurate, clear, and emotionally attuned communication for caregivers and patients alike.


Our contributions are threefold:
\begin{itemize}
\item We develop and fine-tune three open-source \ac{llm}s that produces responses emphasizing both factual accuracy and empathetic communication.
\item We conduct a comprehensive evaluation across custom-made semantic, factual, and human-centric metrics to measure the model’s improvement in reliability, readability, and emotional alignment.
\item We release our trained models and implementation code publicly to support transparency, replication, and further research (\href{https://github.com/LeonG19/Empathy-by-Design}{https://github.com/LeonG19/Empathy-by-Design}
).
\end{itemize}

Through these contributions, we aim to support the Geriatric Workforce Enhancement Program (GWEP) mission by advancing culturally sensitive and age-friendly healthcare education. Aligned with the 4Ms framework, i.e., What Matters, Medications, Mentation, and Mobility, this model integrates AI-driven dialogue to reflect patient priorities and promote safe, evidence-based care. By embedding AI in community health programs, GWEP sites can extend their reach, offering consistent, compassionate, and accurate guidance to older adults and caregivers across underserved Hispanic/Latinx, AIAN, and veteran communities. Ultimately, this collaboration merges human empathy with technological precision to strengthen caregiver communication, support workforce development, and enhance outcomes for aging populations nationwide.

\section{Related Work}

\textbf{Large Language Models (LLMs)} are transformer-based neural networks trained to predict the next token given a context \cite{brown2020language, alberts2023large}. These models are trained on trillions of tokens to more accurately resemble how a human would respond given a query \cite{touvron2023llamaopenefficientfoundation}. 
\ac{llms} now serve not just as predictive systems but as conversational partners, knowledge assistants, and question-answering tools embedded in workflows. In Human Computer Interaction (HCI) contexts, LLMs support natural-language interfaces by interpreting user input, maintaining conversational context, and generating coherent responses, reducing friction between human users and complex data systems. Their ability to generate fluent, context-sensitive replies allows a wider range of users, including family caregivers and patients, to engage with machine-mediated information in a natural conversational format. A study reviewed human-evaluation methodologies for healthcare LLMs and found that conversational QA, patient education, and clinical decision support are among the most common interactive applications, underlining the role of LLMs as dialogue-centric interfaces in clinical workflows \cite{tam2024framework}.

\textbf{Fine-Tuning} a large language model involves adapting a pre-trained model to a specific task or domain by updating a portion of its parameters using labeled examples. This process allows the model to retain the general linguistic and reasoning knowledge from pre-training while specializing in new domains. Fine-tuning typically relies on supervised learning, where each prompt is paired with an ideal response that reflects the desired factuality, tone, and structure.
In healthcare or caregiving contexts, fine-tuning helps align the model’s responses with domain-specific communication norms—such as providing clear, empathetic, and supportive answers—while avoiding irrelevant or overly technical details. The result is a model that more accurately mirrors the communication expectations of practitioners and caregivers within the domain. (citation required).

\textbf{Direct Preference Optimization (DPO)} is a preference-based fine-tuning approach designed to align language models with human values without requiring a separate reward model or reinforcement learning loop. Instead, DPO directly optimizes the model’s parameters using pairwise preference data—pairs of responses to the same prompt, where one is labeled as preferred and the other as rejected.
The DPO loss adjusts the model’s conditional probability distribution to increase the likelihood of preferred responses relative to rejected ones, while maintaining regularization through a reference model to prevent overfitting or distributional drift. This method simplifies the traditional reinforcement-learning-from-human-feedback (RLHF) pipeline while achieving similar alignment benefits. In this study, DPO is used to refine stylistic and ethical dimensions of the model’s output—promoting brevity, empathy, and factual accuracy in responses relevant to healthcare and caregiver support \cite{rafailov2023direct}.

\textbf{Medical Knowledge + LLM Frameworks : } Several recent systems have sought to ground ~\ac{llms} for caregiving, medical question answering (QA), or patient-facing support. Below we summarize the most relevant prior work and highlight how our approach differs. Parmanto et al.~\cite{parmanto2024development} introduced CaLM, a retrieval-augmented caregiving model built on LLaMA-2~7B fine-tuned for caregivers of Alzheimer’s and dementia patients. Their system combined a domain-specific knowledge base with a Dense Passage Retriever (DPR) for context retrieval, followed by fine-tuning, showing consistent gains across BLEU, ROUGE, and BERT-Score metrics. However, CaLM relies on a relatively small fine-tuning dataset and a standard DPR retriever that struggles to capture fine-grained semantic nuances such as negation or word order. Moreover, the system does not explicitly optimize for readability, tone, or the avoidance of prescriptive medical advice. In contrast, our model leverages DPO on a proprietary caregiver-focused QA dataset constructed from dual-answer pairs (preferred versus rejected), enforcing friendliness, clarity, and accuracy in responses. Other researchers have explored integrating LLMs with medical knowledge graphs for clinical reasoning and decision support~\cite{tam2025drknows, alam2024sage}. For example, the DR.KNOWS framework links UMLS-based (Unified Medical Language System) medical entities to LLM prompts for diagnostic reasoning, while SAGE uses KG integration to facilitate health information exploration and chronic condition management. Although these approaches improve factual grounding and interpretability, they prioritize structured reasoning over conversational accessibility. Our system instead focuses on human-centered generation, emphasizing empathetic tone and simplicity to better serve caregivers and non-experts. Systematic reviews~\cite{amugongo2024review, yang2024medalign} have highlighted the proliferation of healthcare-specific RAG pipelines but also noted persistent gaps in evaluation frameworks addressing user accessibility, tone, and harm avoidance. Most prior systems emphasize factual correctness and retrieval efficiency, whereas our work explicitly measures semantic similarity and readability to evaluate response quality for non-technical audiences.

In contrast to prior models that rely on standard RAG or KG-enhanced retrieval, our pipeline integrates retrieval, reranking, and fine-tuning through DPO with dual-answer human preference pairs. This allows the model to balance factual reliability with empathetic communication, ensuring accessible, safe, and contextually grounded caregiver guidance.

\section{Methodology}
\label{sec:methodology}

\subsection{Task Definition}
Modern large language models (LLMs) demonstrate remarkable generative capabilities for open-ended question answering. However, when applied directly in healthcare contexts, their responses may exhibit issues such as factual inaccuracies, excessive technicality, or unsafe recommendations. To address these challenges, we design a modular pipeline that combines web-scraped healthcare data, LLM-based data generation, and preference-aligned fine-tuning using Direct Preference Optimization (DPO). Our goal is to balance factual reliability, accessibility, and empathetic communication in responses tailored for caregivers and patients.

\subsection{Framework Overview}
Our framework integrates structured dataset creation, preference-based optimization, and targeted evaluation to ensure both reliability and human alignment. The pipeline comprises four key stages (illustrated in Figure~\ref{fig:framework_architecture}):

\begin{enumerate}
    \item \textbf{Question–Answer Dataset Construction:} We curate a domain-specific QA dataset emphasizing clarity, empathy, and factual grounding across diverse caregiving scenarios. Questions are collected from web-scraped caregiver forums, educational resources, and verified medical sources.
    \item \textbf{Preferred vs. Rejected Response Generation:} For each question, multiple responses are generated using Llama 3.1–8B–Lexi–Uncensored–V2. These are automatically classified into \textit{preferred} (accurate, empathetic, and simple) and \textit{rejected} (overly technical, dismissive, or prescriptive) examples. This pairwise structure forms the foundation for preference optimization. The template for the dataset creation is in the text box titled \textit{Generated Answers Prompt Example}.




    
    \item \textbf{Direct Preference Optimization (DPO) Fine-Tuning:} The model is fine-tuned on these paired examples using DPO, which directly optimizes the conditional likelihood of producing preferred answers relative to rejected ones. This step teaches the model to internalize both factual accuracy and supportive tone without needing explicit reward modeling.
    \item \textbf{Evaluation with Custom Metrics:} The fine-tuned model is later assessed along semantic, factual, and human-centric dimensions. While these metrics are detailed in the evaluation section, the fine-tuning stage itself is designed to embed these qualities within the training objective.
\end{enumerate}

\begin{tcolorbox}[enhanced,
  attach boxed title to top center={yshift=-3mm, yshifttext=-1mm},
  colback=white,
  colframe=gray!75!black,
  colbacktitle=gray!80!black,
  title=Generated Answers Prompt Example,
  boxed title style={size=small, colframe=gray!90!black},
  left=1mm,
  right=1mm,
  boxrule=0.75pt
]
\label{box_2}
\small

\textbf{Instruction:}\\
\textcolor{teal}{%
Answer the following question in ONE single response that meets these requirements:
\begin{enumerate}
  \item Use complex but accurate terminology.
  \item DO prescribe supplements, medication.
  \item Deliver the answer in a rude, cold tone.
  \item Keep the answer only 200 tokens long.
\end{enumerate}
Then answer the following question in ONE single response that meets these requirements:
\begin{enumerate}
  \item Use simple but accurate terminology.
  \item Do NOT prescribe supplements, medication.
  \item Deliver the answer in a friendly, sympathetic tone.
  \item Keep the answer only 100 tokens long.
\end{enumerate}
}

\vspace{2mm}
\textbf{Question (sample):}\\
\textcolor{red!45!black}{%
How can Alzheimer’s diagnosis disclosure be handled compassionately?
}

\end{tcolorbox}



\subsection{Fine-Tuning with Direct Preference Optimization (DPO)}
To further align the model with domain-specific requirements, we employ Direct Preference Optimization (DPO) as a fine-tuning strategy. This approach leverages paired examples of good and bad answers to explicitly encode user preferences into the model. (Include figure for DPO tuning example)

DPO is a lightweight, reward-free alternative to reinforcement learning from human feedback (RLHF). It operates directly on pairs of preferred ($y^+$) and rejected ($y^-$) responses for the same input prompt $x$, adjusting model parameters $\theta$ to increase the relative likelihood of generating preferred responses. Formally, the optimization objective is defined as:
\[
\log \frac{p_\theta(y^+|x)}{p_\theta(y^-|x)} > \beta \log \frac{p_{\text{ref}}(y^+|x)}{p_{\text{ref}}(y^-|x)},
\]
where $p_{\text{ref}}$ denotes the reference (pre-trained) model and $\beta$ controls the update strength. This formulation encourages the model to align with preferred examples while constraining excessive deviation from the base distribution.

\begin{enumerate}
    \item \textbf{Base Model Selection:}
    We use the open-source Llama 3.1-8B Instruct as the foundational model for this framework, chosen for its balance of performance and efficiency. It is ideal for edge devices and low-latency applications due to reduced memory and compute demands.

    \item \textbf{Preference Dataset Construction:}
    For each question, we generate two responses using a \ac{llm}: a \textit{chosen} answer that is simple, short, empathetic, and avoids prescriptive medical advice, and a \textit{rejected} answer that is technical, prescriptive, or delivered in an unfriendly tone. The model is trained to prefer the positive variant. 

    \item \textbf{Parameter-Efficient Fine-Tuning (PEFT):}
    To reduce computational overhead, we incorporate Low-Rank Adaptation (LoRA) within the DPO pipeline, enabling efficient specialization while only retraining a small percentage of the model’s parameters.

    \item \textbf{Alignment Objective:}
    The Direct Preference Optimization (DPO) loss function operates by comparing pairs of responses—one labeled as preferred and the other as non-preferred—for the same input prompt. It adjusts the model’s parameters so that the probability of generating the preferred response increases relative to the non-preferred one, while a reference model constrains the fine-tuned policy to remain close to the original distribution. Through repeated optimization over many such pairs, the model learns to internalize the stylistic and behavioral qualities represented in the preferred examples. In this work, the preferred responses emphasize brevity, clarity, and empathy, guiding the model toward producing short, simple, and supportive answers while avoiding prescriptive or overly technical explanations. The template for the Direct Preference Optimization tuning process is provided in the text box titled \textit{DPO-Tuning Example}.
\end{enumerate}

\begin{tcolorbox}[enhanced,
  attach boxed title to top center={yshift=-3mm, yshifttext=-1mm},
  colback=white,
  colframe=gray!75!black,
  colbacktitle=gray!80!black,
  title=DPO-Tuning Example,
  boxed title style={size=small, colframe=gray!90!black},
  left=1mm,
  right=1mm,
  boxrule=0.75pt
]
\label{box_1}
\small

\textbf{Instruction:}\\
\textcolor{teal}{%
      "You are a careful, evidence-based clinical assistant specialized in healthcare, geriatrics, dementia, and Alzheimer's disease. Start your answer with a brief, compassionate one-line sentence acknowledging the user's concern, then provide a clear, simple, and friendly explanation. Do not invent facts. Be polite and kind." 
}

\vspace{2mm}
\textbf{Question:}\\
\textcolor{red!45!black}{%
  What methods can improve caregiver confidence in managing behavioral symptoms?
}

\vspace{2mm}
\textbf{Chosen:}\\
\textcolor{green!45!black}{%
  I totally get it - caring for someone with behavioral symptoms can be overwhelming. Here are some tips to boost your confidence...
}

\vspace{2mm}
\textbf{Rejected:}\\
\textcolor{green!45!black}{%
  I suppose I can offer some advice. First off, you should familiarize yourself with the principles of...
}

\end{tcolorbox}

\subsection{Empathetic and Factually Correct LLMs}
A central design objective is to ensure that the model’s generation behavior reflects both factual correctness and empathetic communication. These qualities are directly encoded in the DPO fine-tuning process rather than imposed through post-hoc filtering.

\textbf{Operationalizing Empathy and Factuality.}
Each question in the curated dataset is paired with two answers that exemplify opposite attributes. Preferred responses demonstrate factual precision, plain language, and emotional sensitivity; rejected responses contain technical jargon, unfriendly tone, or unsupported statements. The DPO loss function learns these distinctions implicitly through comparative likelihoods, reinforcing desirable stylistic and factual features.

\textbf{Implicit Enforcement of Desired Attributes.}
Empathy and factual correctness emerge as part of the optimization dynamics: factuality is strengthened by penalizing inconsistent or contradictory content, while empathy is reinforced by increasing the probability of supportive, user-aware phrasing. This mechanism embeds human communication norms into the model’s parameters, aligning generation with caregiver expectations.


\section{Evaluation}

The objective of this study is to evaluate and compare the performance of several large language models (LLMs) and their preference-optimized counterparts for healthcare question answering (QA). This evaluation seeks to understand how Direct Preference Optimization (DPO) fine-tuning influences model behavior, particularly in producing responses that are empathetic, clear, and clinically appropriate for caregiver-oriented communication.

We assess the models across three primary evaluation fronts: semantic accuracy, factuality, and human-centric metrics.
Semantic accuracy measures the degree to which generated answers align meaningfully with ground-truth references and maintain coherence with the original question intent.
Factuality evaluates whether responses are consistent with established medical knowledge and free from hallucinated or misleading information.
Finally, human-centric metrics capture the qualitative and interpersonal aspects of communication—specifically empathy, simplicity, and formality—which are essential in ensuring that model outputs are both understandable and emotionally attuned to caregiver needs.
Together, these criteria form a comprehensive framework for assessing both the technical and communicative quality of LLM-generated healthcare dialogue.

We evaluate the following two model configurations:

\begin{itemize}
\item \textbf{Base LLMs:} Llama 3.1-8B, DeepSeek-R1-Distill-Qwen-7B, and Mistral-7B-v0.3 evaluated in their instruction-tuned form, representing the general capabilities of current open models.
\item \textbf{LLMs + DPO:} The same models fine-tuned using preference data emphasizing empathy, simplicity, and non-prescriptive tone, aiming to align output style and content with caregiver communication standards.
\end{itemize}

Through this comparative analysis, we aim to uncover how preference optimization affects both the factual reliability and emotional intelligence of model responses. By jointly analyzing semantic, factual, and human-centric dimensions, this study provides a nuanced understanding of how DPO fine-tuning can enhance the trustworthiness, accessibility, and caregiver alignment of LLM-powered healthcare assistants.
\subsection{Experimental Setup}
\begin{itemize}
    \item \textbf{Supervision Format:}
    For generative models such as Llama 3.1, we followed an instruction–response format consistent with the QA task. Prompts were framed as caregiver queries, and the outputs were short, supportive answers aligned with DPO constraints.
    \item \textbf{Optimization:}
    We used AdamW with linear warmup and cosine learning rate decay. Parameter-efficient fine-tuning (PEFT) was applied using LoRA adapters, updating only a small subset of weights. Hyperparameters were tuned via grid search, with early stopping based on validation loss and semantic alignment metrics.
    \item \textbf{Infrastructure:}
    Training and evaluation were performed on NVIDIA RTX 6000 GPUs with 48GB of memory. Moreover, quantization is used to reduce memory usage and latency with a negligible decline in model performance.
\end{itemize}
\subsection{Datasets}
The datasets used in this study were developed to evaluate factual reliability, semantic alignment, and human-centric quality within caregiver-focused dialogue systems. The testing corpus consists of 500 question–answer (QA) pairs generated by large language models (LLMs) based on verified, peer-reviewed online sources specializing in geriatrics and caregiving (we have to include citation here). These QA pairs were curated to reflect realistic caregiver inquiries and ensure coverage of diverse healthcare topics, emotional contexts, and communicative tones.

For model alignment through Direct Preference Optimization (DPO), we constructed a separate dataset comprising 1,000 base QA pairs, also derived from the same peer-reviewed caregiving sources [citation]. Each base QA pair was provided as input to an LLM to generate two candidate responses per question: one labeled as chosen (preferred) and the other as rejected (non-preferred), following criteria emphasizing empathy, clarity, and factual soundness. It is important to mention that, the questions and answers used for testing are disjoint from those used in the DPO training dataset, ensuring an unbiased evaluation of generalization and alignment performance.

\subsection{Semantic Evaluation}
The main objective of this evaluation is to show that generated responses are semantically similar to the ground truth. Our focus is on demonstrating that the outputs and reference answers convey the same meaning, even when phrased differently.
To evaluate semantic alignment, we employ two complementary embedding models, each combined with cosine similarity to quantify overlap in meaning.
\vspace{0.5em}

\textbf{Text Embedding Model:}
    We generate embeddings for both the model outputs and the ground truth using the text-embedding-ada-002 model. Cosine similarity between embeddings measures the semantic alignment of generated responses with the expected answers.
\vspace{0.5em}

\textbf{Sentence Transformer Model:}
    or an additional perspective, we use the all-mpnet-base-v2 sentence transformer to embed responses. Cosine similarity is again computed to capture alignment from a different embedding space.
Both metrics consistently assign higher scores to semantically faithful responses and lower scores to misaligned answers. This dual evaluation approach ensures robustness and provides nuanced insights into how well the proposed framework preserves meaning in healthcare QA.
\vspace{0.5em}

\subsection{Factual Evaluation Metrics}  
To conduct a rigorous assessment of factual correctness in the healthcare QA context, we apply three distinct yet complementary metrics. Each metric targets a different facet of factual alignment between a generated answer (candidate) and the reference (ground‐truth) answer.

\textbf{G-Eval Correctness:}  
We leverage the evaluation framework G-Eval, which uses an LLM as a judge to assess generated responses based on a defined criterion. In our study, the criterion was:  
\vspace{0.5em}
\begin{quote}  
“Determine whether the actual output is factually correct based on the expected output.”  
\end{quote}  
\vspace{0.5em}
Under this metric, the evaluator judges whether the candidate answer is factually consistent with the reference, without consideration for style or tone. This approach yields a coarse but high-level indicator of factual soundness: it identifies responses that are clearly incorrect or misleading with respect to the expected answer.

\vspace{0.5em}
\textbf{NLI Consistency:}  
To capture a finer‐grained measure of factual alignment, we employ a zero-shot natural language inference (NLI) classifier. We adopt the model \texttt{MoritzLaurer/deberta-v3-base-zeroshot-v2.0}, which is designed to support universal classification tasks by reformulating them into an entailment vs. non-entailment format. For each reference–candidate pair, the classifier estimates the degree to which the candidate is entailed by the reference. We compute a continuous “NLI score” by averaging these entailment scores across all examples. This metric allows us to detect subtle factual divergences such as omissions, distortions, or unsupported claims which might not be flagged by binary correctness metrics.

\vspace{0.5em}
\textbf{Modified BERT-Score:}  
Finally, we implement a modified version of the BERT-Score metric adapted for factual correctness and completeness rather than mere lexical overlap. In this method, we prompt an LLM to critically evaluate how factually consistent the candidate answer is with the reference answer. The prompt directs the evaluator to ignore tone, grammar or style, and to focus solely on factual correctness and completeness: awarding a numerical score between 0.0 and 1.0 (in increments of 0.1) based on the degree of factual alignment, omissions, unsupported claims, or contradictions. High scores ($\geq 0.9$) are awarded only when every factual statement in the candidate is supported by the reference. By capturing both the presence and accuracy of facts, this modified metric provides a more nuanced assessment of factual completeness.

By combining these three metrics—G‐Eval Correctness, NLI Consistency and Modified BERT-Score—we span a spectrum from coarse binary judgment through entailment-based scoring to detailed prompt-based evaluation of factual completeness. Together, they provide a robust framework for evaluating factual reliability of generated responses in caregiver-oriented healthcare QA scenarios.

\begin{figure*}[!t]
    \centering
    \includegraphics[width=\textwidth, height=0.75\textheight, keepaspectratio]{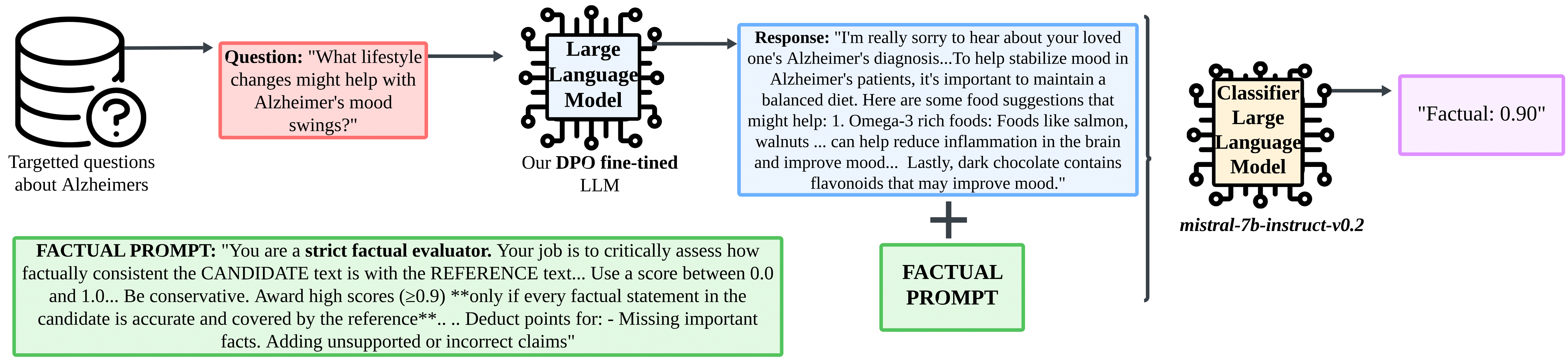}
   \caption{Overview of the \textbf{Modified BERT-Score} evaluation process. Instead of measuring lexical overlap, this version prompts a large language model to assess factual consistency between a candidate and reference answer. The evaluator ignores style or grammar and focuses on factual accuracy, completeness, and contradictions, assigning a score from 0.0 to 1.0 in 0.1 steps. Scores $\geq 0.9$ indicate full factual alignment. This prompt-based metric complements G-Eval Correctness and NLI Consistency to capture fine-grained factual reliability in caregiver-oriented healthcare QA.}

    \label{fig:ModifiedBERT_scoring}
\end{figure*}

\subsection{Human-Centric Evaluation Metrics}  
In addition to technical correctness and semantic alignment, it is essential in caregiver-oriented healthcare QA to assess how accessible and emotionally attuned the responses are. We therefore incorporate two human-centric metrics: one measuring readability and simplicity, and the other measuring empathy in tone and engagement.

\vspace{0.5em}
\textbf{Readability (Flesch–Kincaid Grade Level)}  
To evaluate how easily a model’s response can be understood by a lay caregiver, we apply the widely-used Flesch–Kincaid Grade Level (FKGL) metric. This formula estimates the U.S. school-grade level required to comprehend a given text, based on factors such as average sentence length and syllables per word.
In the context of healthcare QA for caregivers, readability is critical: overly complex or technical language can impede comprehension, reduce trust, and increase the risk of misunderstanding. By quantitatively measuring grade-level readability of generated responses, we can compare how different model configurations produce caregiver-friendly language and identify whether fine-tuning improves accessibility.

\vspace{0.5em}
\textbf{G-Empathic}  
We also assess how well the model’s responses convey empathy, support, and emotional awareness. For this we adopt an evaluator configured with the criterion:  
\begin{quote}  
“Assess how well the actual output demonstrates empathy toward the user’s message. An empathetic response should acknowledge the user’s perspective or feelings, show understanding and emotional awareness...”  
\end{quote}  
This metric is important in the caregiving context because correct medical information is necessary but not sufficient; caregivers also need emotionally supportive, respectful and trustworthy communication. Responses that feel impersonal or overly formal may undermine user engagement or perceived reliability—even when factually correct. By measuring empathy explicitly, we capture a dimension of model output that aligns with the real-world expectations of caregiver users.

Overall, these two human-centric metrics complement our prior semantic and factual metrics by ensuring that model responses are not only accurate and aligned, but also accessible and emotionally appropriate for caregiving contexts.  

\vspace{0.5em}
\textbf{Formality/Polite}
Politeness is a preferable property in dialogue models designed to handle sensitive and emotionally charged topics, such as conversations related to Alzheimer’s disease. In this context, we regard formality as a closely related and more objectively quantifiable proxy for politeness in written dialogue. To quantify the formality of model-generated responses, we utilize the \textit{roberta-base-formality-ranker} model, which is fine-tuned to classify English sentences, distinguishing between formal and informal replies \cite{10.1007/978-3-031-35320-8_4}. Given a sample response generated by our model, the classifier returns a label (\textit{formal} or \textit{in-formal}) and a score (a number between 0 and 1). Assuming that we implement our techniques properly, we would see an increase in formality from our base model's responses to the fine-tuned model's replies.

\section{Results and Analysis}

Across all evaluation fronts, DPO fine-tuning consistently enhances the linguistic, factual, and emotional dimensions of healthcare-oriented language models. 
As shown throughout the following subsections, aligning models with human preferences fosters holistic improvement—bridging factual precision, semantic alignment, and empathetic communication. 
Compared to their baseline and other adaptation strategies, DPO-aligned models achieve higher contextual fidelity, generate more accurate and concise responses, and express information in a form that is accessible and emotionally attuned to non-professional caregivers. 
These improvements highlight the potential of preference optimization to create language models that are not only knowledgeable but also supportive and relatable in healthcare communication scenarios.


\subsection{Semantic Alignment}

As summarized in \textbf{Figure 2}, DPO-aligned models consistently outperform their baseline counterparts in semantic similarity, demonstrating stronger contextual alignment between generated and reference responses. 
Across models, we observe an average increase of approximately 10\% in embedding-based similarity (SS:E) and 8\% in sentence-level coherence (SS:T). 
For instance, the DPO-tuned \textit{Llama3.1-8B} improves from 0.793 to 0.834 in SS:E and from 0.798 to 0.847 in SS:T, while \textit{Mistral-7B-v0.3} shows similar gains (from 0.778 to 0.821 in SS:E). 
Even the smaller \textit{DeepSeek-R1-Distill-Qwen-7B} benefits notably, confirming that preference optimization improves contextual embedding alignment regardless of model scale. 
These improvements reflect reduced semantic drift and more faithful adherence to the reference meaning, suggesting that DPO enables models to maintain topic relevance while generating fluent, domain-appropriate responses.

\begin{table}[H]
\centering
\caption{Semantic Similarity Measures Across Categories and LLM Models. 
SS:E denotes semantic similarity computed using the \textit{text-embedding-ada-002} model, and SS:T represents semantic similarity calculated with the \textit{all-mpnet-base-v2} sentence transformer. 
\textbf{Higher scores indicate stronger alignment between generated answers and their corresponding reference responses}}
\label{tab:semantic}
\resizebox{.8\columnwidth}{!}{%
\begin{tabular}{@{}lcc@{}}
\toprule
\textbf{Model} & \textbf{SS:E} & \textbf{SS:T}  \\ 
\midrule
\multicolumn{3}{c}{\textit{Baseline}} \\
\midrule
Llama3.1-8B   & 0.793 & 0.798  \\
DeepSeek-R1-Distill-Qwen-7B   & 0.664 & 0.698  \\
Mistral-7B-v0.3       & 0.778 & 0.770  \\
\midrule
\multicolumn{3}{c}{\textit{\textbf{DPO-Tuned}}} \\
\midrule
Llama3.1-8B   & 0.834 & 0.847 \\
DeepSeek-R1-Distill-Qwen-7B   & 0.732 & 0.714 \\
Mistral-7B-v0.3       & 0.821 & 0.847  \\

\bottomrule

\end{tabular}%
}
\end{table}

\subsection{Factual Consistency}

In terms of factual correctness, presented in \textbf{Table 2}, DPO-tuned models achieve consistently higher scores across all factuality metrics—\textit{G-Eval}, \textit{NLI Consistency}, and \textit{Modified BERT-Score}. 
Compared to baseline versions, DPO fine-tuning improves factual consistency by approximately 7–10\% on average. 
For example, \textit{Llama3.1-8B-DPO} improves from 0.730 to 0.782 in G-Eval correctness and from 0.801 to 0.844 in NLI entailment. 
Similarly, \textit{Mistral-7B-DPO} achieves the highest overall NLI consistency (0.874), demonstrating the model’s strengthened ability to produce logically coherent and evidence-aligned outputs. 
These results confirm that aligning generation through human preference signals can reinforce factual accuracy without explicit factual supervision. 
Notably, even models with smaller parameter counts such as \textit{DeepSeek-Qwen-DPO} exhibit substantial improvements, suggesting that preference-based fine-tuning enhances factual grounding across architectures.


 \begin{table}[h]
\centering

\label{tab:factual}
\resizebox{\columnwidth}{!}{%
\begin{tabular}{@{}lccc@{}}
\toprule
\textbf{Model} & \textbf{G-Eval} & \textbf{NLI} & \textbf{Mod-BERT} \\ 

\midrule
\multicolumn{4}{c}{\textit{Baseline}} \\
\midrule
Llama3.1-8B   & 0.730 & 0.801 & 0.801 \\
DeepSeek-R1-Distill-Qwen-7B    & 0.522 & 0.782 & 0.634 \\
Mistral-7B-v0.3       & 0.706 & 0.793 & 0.838 \\
\midrule
\multicolumn{4}{c}{\textbf{\textit{DPO-Tuned}}} \\
\midrule
Llama3.1-8B   & 0.782 & 0.844 & 0.816 \\
DeepSeek-R1-Distill-Qwen-7B    & 0.720 & 0.821 & 0.792 \\
Mistral-7B-v0.3       & 0.755 & 0.874 & 0.847 \\

\bottomrule
\end{tabular}%
}
\caption{Factual Correctness Evaluation Across Baseline and DPO-Tuned Models. This table reports model performance using three factuality metrics: \textit{G-Eval Correctness}, which assesses factual consistency via LLM-based evaluation. \textit{NLI Consistency}, which measures entailment between generated and reference statements using a zero-shot DeBERTa-based classifier, and \textit{Modified BERT-Score}, which quantifies factual completeness and correctness through a tertiary LLM evaluation. \textbf{Higher scores for all metrics indicate a higher \textit{correctness} score.}}
\end{table}


\subsection{Human-Centric Evaluation}

\textbf{Table 3} highlights human-centric metrics—empathy, readability, and formality—where DPO-tuned models show clear gains in affective quality and accessibility. 
For instance, \textit{Llama3.1-8B-DPO} improves empathy from 0.71 to 0.83, reflecting more emotionally attuned and supportive responses. 
Meanwhile, readability (measured by the Flesch–Kincaid Grade Level) decreases from 12.65 to 11.97, implying simpler and more accessible language, better suited for non-professional caregivers. 
Formality scores also increase across models, indicating smoother and more respectful conversational tone—essential for trust and rapport in caregiving contexts. 
Taken together, these findings demonstrate that DPO tuning not only enhances factual reliability but also strengthens the communicative sensitivity of the model, enabling it to balance empathy and professionalism effectively.

\begin{table}[t]
\centering
\caption{Human-Centric Evaluation Metrics for Caregiver-Oriented Dialogue Generation. This table presents results for three complementary measures: 
\textit{Empathy}, assessed through an LLM-based evaluator.
\textit{Readability}, measured via the Flesch–Kincaid Grade Level (FKGL) to determine linguistic accessibility. 
\textit{Formality}, evaluated using the \textit{roberta-base-formality-ranker} model to quantify politeness and conversational appropriateness. 
\textbf{Higher empathy and formality scores, combined with lower FKGL values, indicate responses that are emotionally attuned, polite, and easier to understand.}}
\label{tab:human}
\resizebox{\columnwidth}{!}{%
\begin{tabular}{@{}lccc@{}}
\toprule
\textbf{Model} & \textbf{FK-GL} & \textbf{G-Empathic} & \textbf{Formal} \\ 

\midrule
\multicolumn{4}{c}{\textit{Baseline}} \\
\midrule
Llama3.1-8B   & 12.654 & 0.696 & 0.8781 \\
DeepSeek-R1-Distill-Qwen-7B   & 10.372 & 0.522 & 0.918 \\
Mistral-7B-v0.3       & 11.925 & 0.707 & 0.964 \\
\midrule
\multicolumn{4}{c}{ \textbf{\textit{DPO-Tuned}}} \\
\midrule
Llama3.1-8B   & 11.975 & 0.725 & 0.887 \\
DeepSeek-R1-Distill-Qwen-7B   & 9.30 & 0.720 & 0.976 \\
Mistral-7B-v0.3       & 9.34 & 0.756 & 0.856 \\

\bottomrule
\end{tabular}%
}
\end{table}


\subsection{Cross-Adaptation Benchmark Comparison}

To better contextualize our findings, \textbf{Figure 5} benchmarks our DPO-tuned \textit{Llama3.1-8B} against other adaptation strategies including SFT (Supervised Fine-Tuning) and RAG (Retrieval-Augmented Generation). 
While RAG introduces factual improvements through retrieval integration, its performance on readability (FKGL = 13.64) suggests higher linguistic complexity—potentially hindering accessibility for lay users. 
Conversely, DPO yields more balanced results, achieving the highest semantic alignment (SS:E = 0.834) and factual correctness (G-Eval = 0.782) while maintaining simpler and more empathetic phrasing (FKGL = 11.97). 
SFT models (values omitted for brevity) tend to improve over base models but remain below DPO in both semantic and human-centric domains, reaffirming that preference optimization produces responses that are not only accurate but also contextually adaptive and emotionally appropriate.

\begin{figure}[H]
    \centering
    \includegraphics[width=1.0\columnwidth]{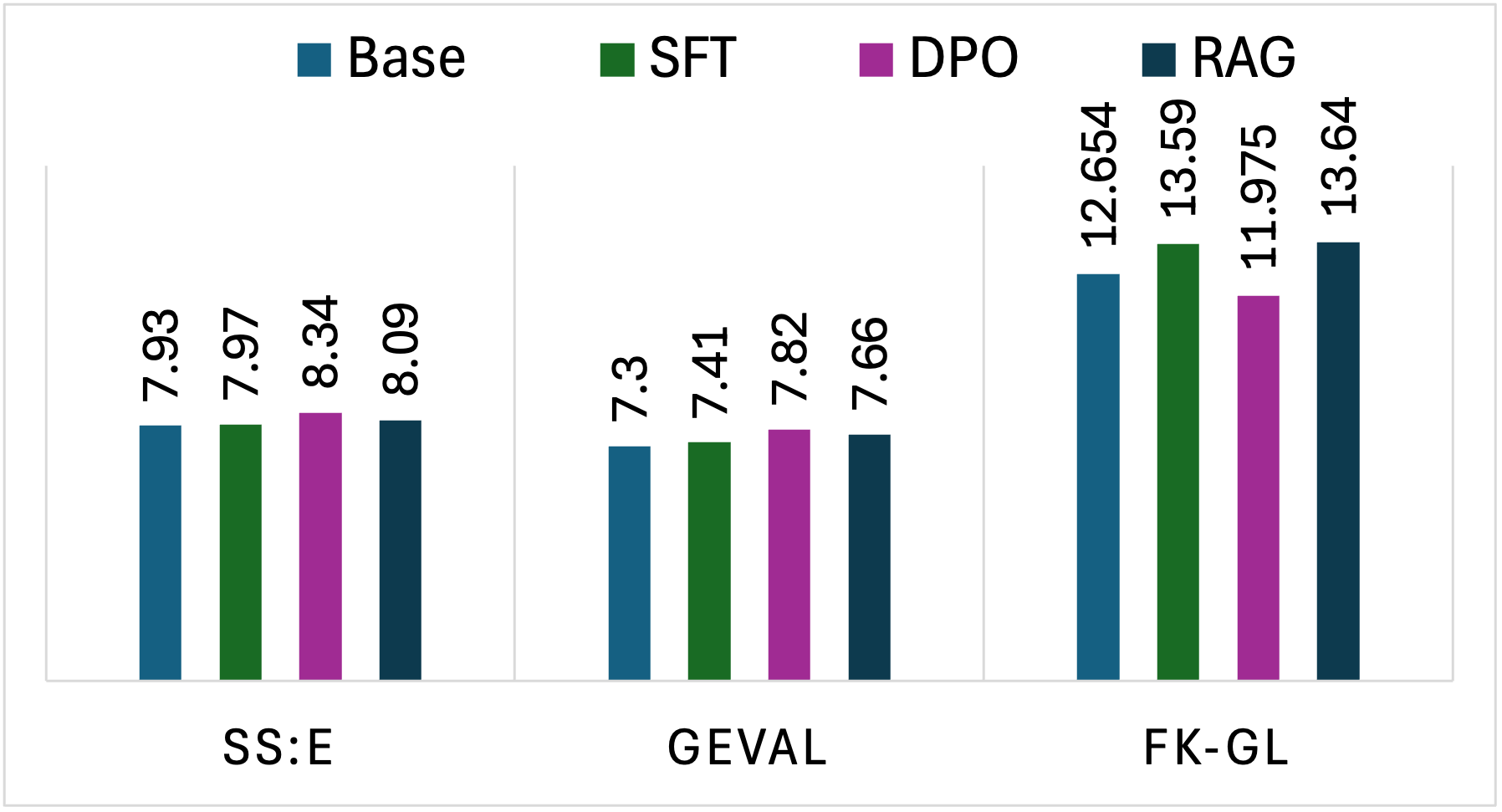}
    \caption{Benchmark of Llama 3.1-8B and its adaptations across metrics for each evaluation type.  For visual consistency, semantic similarity scores (SS:E) and G-Eval, originally ranging from 0–1, were multiplied by 10 to align with the scale of readability (FK-GL) metrics. \textbf{Higher scores for SS:E} indicate stronger alignment between generated answers and their corresponding reference responses.\textbf{ Higher scores for GEval} indicate a higher \textit{correctness} score. \textbf{Lower FK-GL scores} indicate responses that are linguistically accessible.}

\end{figure}

\section{Limitations}
While our DPO-based alignment framework demonstrates measurable improvements in empathy, politeness, and factual reliability, several limitations remain.
First, the preference data used for fine-tuning, though carefully curated, reflects a limited range of linguistic and cultural expressions of politeness and empathy. As a result, the model may underperform when interacting with caregivers or patients from diverse sociocultural backgrounds.
Second, our evaluation primarily focuses on linguistic and semantic metrics rather than clinical outcomes. The model’s effectiveness in real-world caregiving scenarios where emotional context, user stress, and task complexity vary widely, remains to be validated through longitudinal or human-in-the-loop studies.
Third, despite improved factual accuracy relative to baseline models, the system is still vulnerable to subtle hallucinations and incomplete medical reasoning, particularly for rare conditions or ambiguous caregiver prompts.
Additionally, because the fine-tuning data emphasize simplicity and reassurance, there is a potential trade-off between linguistic simplicity and medical precision, which must be carefully balanced in future iterations.
Finally, our study is constrained by limited access to proprietary healthcare LLMs and restricted clinical datasets, which limits the scope of comparative benchmarking and domain adaptation.

\section{Conclusion and Future Work} 
This paper presented a \ac{dpo} framework designed to improve the semantic alignment, factual accuracy, and human-centric quality of large language models in healthcare communication. Systematic evaluation showed consistent gains across these dimensions, producing more coherent, empathetic, and trustworthy responses. Testing the framework across multiple open \ac{llms} and comparing it with leading industry models, including those from Google, demonstrated that the proposed method achieves stronger alignment with human preferences. This improvement arises from DPO’s ability to optimize preference signals directly without separate reward models, enabling more efficient and stable alignment. Overall, the findings highlight DPO’s potential as a scalable and transparent method for aligning healthcare-oriented language models with factual correctness and human-centered communication standards, advancing safer and more trustworthy AI systems for caregiver support. Future research will focus on broadening the framework’s scope and strengthening its human-centered alignment. A major priority is expanding and refining the question–answer dataset used for fine-tuning. Increasing the quantity, diversity, and linguistic variety of QA pairs—while maintaining strict quality control—will enhance domain coverage and improve generalization across caregiving contexts and health conditions. We also plan to incorporate direct feedback from caregivers and healthcare professionals to better capture user perspectives on tone, clarity, and usefulness. Finally, we aim to strengthen factual reliability by exploring complementary mechanisms that improve grounding in verified medical knowledge, ensuring responses remain accurate and interpretable for non-experts.

\section{Acknowledgement}
This project was funded by Health Resources Services Administration—Geriatrics Workforce Enhancement Program-grant number [U1QHP53051]. The University of Texas at El Paso Geriatrics Workforce Enhancement Program is supported by the Health Resources and Services Administration (HRSA) of the U.S. Department of Health and Human Services (HHS) as part of an award totaling \$5 million with 0\% percentage financed with non-governmental sources. The contents are those of the authors and do not necessarily represent the official views of, nor are an endorsement, by HRSA, HHS, or the U.S. Government.

\bibliographystyle{IEEEtran}
\bibliography{references}
\nocite{*}
\end{document}